# Can Large Language Models Logically Predict Myocardial Infarction? Evaluation based on UK Biobank Cohort

## Authors
Yuxing Zhi, BS[a, 1]; Yuan Guo, BS[b, 1]; Kai Yuan, PhD[c, 1]; Hesong Wang, MD[c]; Heng Xu, BS[c]; Haina Yao, BS[b]; Albert C Yang, PhD[d, e]; Guangrui Huang, PhD[c, *]; Yuping Duan, PhD[b, f, *]

## Affiliations:
[a] Qi-Huang School of Chinese Medicine, Beijing University of Chinese Medicine, Beijing, China
[b] Center for Applied Mathematics, Tianjin University, Tianjin, China
[c] School of Life Sciences, Beijing University of Chinese Medicine, Beijing, China
[d] Division of Psychiatry, School of Medicine, National Yang-Ming University, Taipei, Taiwan, China
[e] Department of Psychiatry, Taipei Veterans General Hospital, Taipei, Taiwan, China
f School of Mathematical Sciences, Beijing Normal University, Beijing, China

[1] These authors contributed equally to this work.
[*] **Corresponding Authors:**
Yuping Duan: Center for Applied Mathematics, Tianjin University, 92 Weijin Road, Nankai District, Tianjin 300072, China.
Guangrui Huang: School of Life Sciences, Beijing University of Chinese Medicine, No.11 Beisanhuandong Road, Chaoyang District, Beijing 100029, China.
**E-mail addresses:** yuping.duan@tju.edu.cn (Y. Duan), hgr@bucm.edu.cn (G. Huang).

## Abstract
**Background:** Large language models (LLMs) have seen extraordinary advances with applications in clinical decision support. However, high-quality evidence is urgently needed on the potential and limitation of LLMs in providing accurate clinical decisions based on real-world medical data.
**Objective:** To evaluate quantitatively whether universal state-of-the-art LLMs (ChatGPT and GPT-4) can predict the incidence risk of myocardial infarction (MI) with logical inference, and to further make comparison between various models to assess the performance of LLMs comprehensively.
**Methods:** In this retrospective cohort study, 482,310 participants recruited from 2006 to 2010 were initially included in UK Biobank database and later on resampled into a final cohort of 690 participants. For each participant, tabular data of the risk factors of MI were transformed into standardized textual descriptions for ChatGPT recognition. Responses were generated by asking ChatGPT to select a score ranging from 0 to 10 representing the risk. Chain of Thought (CoT) questioning was used to evaluate whether LLMs make prediction logically. The predictive performance of ChatGPT was compared with published medical indices, traditional machine learning models and other large language models.

**Results:** Low discrimination was found in both ChatGPT (AUC 0.62, 95% CI: 0.58~0.66) and GPT-4 (AUC 0.69, 95% CI: 0.65–0.73) in predicting MI incidence. The CoT approach did not yield substantial improvements in the performance of ChatGPT (AUC 0.58, 95% CI: 0.54~0.62) and GPT-4 (AUC 0.60, 95% CI: 0.56~0.65). In comparison with other models, ChatGPT and GPT-4 underperformed traditional machine learning models (AUC: Decision tree, 0.79, 95% CI: 0.76~0.83; Logistic regression, 0.77, 95% CI: 074~0.81 ; Random forest, 0.76, 95% CI: 0.73~0.80; SVM, 0.79, 95% CI 0.76~0.83), but showed similar performance to other LLMs (AUC: Llama, 0.61, 95% CI: 0.56−0.65; MedLlama, 0.62, 95% CI: 0.57−0.66; Mixtral and Meditron, 0.60 (95% CI: 0.56−0.65).

**Conclusions:** Current LLMs are not ready to be applied in clinical medicine fields. Future medical LLMs are suggested to be expert in medical domain knowledge to understand both natural languages and quantified medical data, and further make logical inferences.

## Introduction

The application of large language models (LLMs) in the medical fields has attracted much attention in the past few months. Consisting of more than billions of weights [1], state-of-the-art LLMs represented by ChatGPT and GPT-4 performed well in recommendation and extraction tasks in cardiovascular disease prevention [2], infectious disease surveillance [3] and neoplasm survival prediction [4] due to its advantages in processing natural languages. However, the applicability of these large models in clinical decision support (CDS), especially providing clinical diagnosis, requires in-depth understanding on real-world medical data, not just generalized medical knowledge. Moreover, the accountability and ethics of LLMs are under widespread doubts and debates [5]. Whether these large models are possible to logically provide accurate clinical diagnosis remains unverified. Therefore, high-quality evidence is urgently needed to evaluate the quantified performance of medical LLMs objectively and clarify their limitations comprehensively with well-designed experiments [6].

Myocardial infarction (MI) is an event of myocardial injury resulted from an imbalance between oxygen supply and demand in which early diagnosis and intervention is critical [7]. Models using statistical methods or artificial intelligence (AI) algorithms have been built to estimate the risk of MI based on structured datasets of clinical symptoms, medical history and biomarkers [8]. Comparatively, LLMs, with the ability to comprehend natural languages and thus in-context learning medical domain knowledge, is providing a novel user-friendly way in MI prevention [9]. Here, we evaluated the predictability of ChatGPT and GPT-4 on the risk of MI during health consultations with prompts constructed from the UK Biobank dataset, and validated the logical inference ability utilizing the Chain of Thought (CoT), hopefully to provide evidence for the potential and limitation of LLMs in CDS tasks.

## Methods

### Data Source and Study Population

This is a retrospective cohort study using the UK Biobank database. UK Biobank is a large-scale biomedical database with over 500,000 participants from England, Wales, and Scotland aged between 40 and 69 [10], with approval from North West Multi-Center Research Ethics Committee renewed in 2021. The study followed the Strengthening the Reporting of Observational Studies in Epidemiology (STROBE) reporting guideline [11].

For the current study, 502,468 participants were originally recruited from 2006 to 2010, up until 2018. Participants without MI incidence at baseline were originally included, in which those without missing data on basic information. For participants with MI incidence, only those with at least one typical symptom reported within one year before MI incidence were kept. A total of 482,310 participants remained to make up the primary cohort for this study including 655 participants with MI incidence. Participants were further re-sampled into a final cohort, in which 316 participants recorded with MI incidence and 374 without MI incidence were sampled from the original data randomly due to low incidence and rate limit of OpenAI API (see Supplementary Appendix eMethod 1). All participants from UK Biobank had provided written informed consent under application number 69344.

### Myocardial Infarction Predictors and Outcome Definition

Common risk factors of cardiovascular diseases were selected for modeling [12]. Predictors include patient demographics, common symptoms, personal history and family history, together with past medical history including comorbidities, medications and laboratory examinations (See Supplementary Appendix eMethod 1). MI was defined following the International Classification of Disease (ICD) with ICD-10 code I22 or ICD-9 code 412 [13].

### Prediction Based on ChatGPT

GPT models, represented by ChatGPT and the more advanced GPT-4, have been widely recognized for their ability to generate text of human-like quality on various topics [14]. In medical consultation, ChatGPT provides longer responses to medical health questions compared to doctors with significantly higher quality and empathy [15]. Building upon this, GPT-4 has further enhanced these capabilities, offering more precise and contextually nuanced responses in medical discussions.

We evaluate both ChatGPT and GPT-4 in their predictability on the risk of myocardial infarction (MI) logically with a series of experiments. The corresponding tabular data of the selected relevant risk factors were first transformed into textual descriptions for identification. Specifically, envisioning ChatGPT and GPT-4 as doctors, we request ChatGPT and GPT-4 to select a score ranging from 0 to 10, representing the closest to estimate MI risk (see Supplementary Appendix eMethod 2).

Firstly, we evaluated the general performance of ChatGPT and GPT-4 with prompts given complete information. Secondly, to further validate whether ChatGPT and GPT-4 make predictions logically, CoT prompts were designed where information used for questioning were given in a stepwise order. Group1 models use basic information and symptoms. Group2 models add personal habits to the inquiry. Group3 models further include family history, building progressively more detailed patient profiles. Finally, multiple comparisons were made between ChatGPT and GPT-4 and other models.

**Model Comparison**

We compared ChatGPT and GPT-4 with three classes of models, namely, medical indices, machine learning models and other representative natural language processing (NLP) models (see Supplementary Appendix eMethod 3). The traditional machine learning models used for comparison include Decision Tree, Logistic Regression, Random Forest, and Support Vector Machine (SVM). Transformer is the mainstream architecture in current NLP models on which Bert and GPT series of models were developed. Our study reported on current NLP models, such as the base Transformer [16] and BERT [17] to analyze the accuracy of mainstream NLP models in medical diagnosis. In addition, we expanded our study to include open-source large language models (LLMs) such as Llama [18] and Mixtral [19]. We also conducted extensive evaluations of models fine-tuned on medical-specific corpora, including MedLlama [18] and Meditron [20], which were instruction-tuned for medical applications. We fine-tuned the aforementioned NLP models on the same textual data posed to GPTs to provide predictions for myocardial infarction in the interviewees, and compare with ChatGPT and GPT-4. Furthermore, indices built by Jee [21] and Wilson [22] were adopted as baseline evaluation model.

**Statistical Analysis**

Continuous variables were expressed as median (IQR), and categorical variables were expressed as numbers (percentage). For each model, receiver operating characteristic (ROC) curves and area under curve (AUC) were used to evaluate model discrimination. Accuracy, precision, sensitivity and specificity were estimated based on confusion matrix at classification threshold selected by maximization of Youden Index, representing the predictive performance of the model. The statistical tests were two-tailed with α=0.05 regarded as statistically significance. Analyses were conducted with R, version 4.1.2 (The R Foundation for Statistical Computing) and PyTorch, version 1.10.0 (The Linux Foundation) performed on a server with 4 NVIDIA RTX 3090 GPUs.

## Results

**Characteristics of Study Population**

Among the primary cohort of 132,008 individuals in this study, 655 participants occurred with MI during follow-up. Descriptive statistics is shown found in Table 1.

Table 1. Baseline characteristics of the study population.[a]

| Characteristics | MI (*n*=316) | Non-MI (*n*=374) |
|---|---|---|
| **Age at recruitment, years** | | |
| mean (SD) | 60 (7) | 57 (7) |
| median (IQR) | 62 (10) | 59 (12) |
| **Sex, n (%)** | | |
| Male | 228 (72.2%) | 173 (46.3%) |
| Female | 88 (27.8%) | 201 (53.7%) |
| **Ethnicity, n (%)** | | |
| White | 284 (90.4%) | 333 (89.3%) |
| Non-white | 30 (9.6%) | 40 (10.7%) |
| **Common symptoms, n (%)** | | |
| Chest pain or discomfort when standing still | 49 (15.5%) | 12 (3.2%) |
| Chest pain or discomfort when walking | 29 (9.2%) | 6 (1.6%) |
| Shortness of breath | 33 (10.4%) | 9 (2.4%) |
| **Comorbidities, n (%)** | | |
| Diabetes mellitus | 37 (11.7%) | 16 (4.3%) |
| Hypertension | 187 (63.6%) | 184 (52.9%) |
| Hypercholesterolemia | 143 (48.3%) | 186 (53.1%) |
| Obesity | 0 (0.0%) | 24 (6.4%) |
| **Medications, n (%)** | | |
| Cholesterol lowering medication | 31 (35.2%) | 28 (14.1%) |
| Blood pressure medication | 34 (38.6%) | 40 (20.1%) |
| Insulin | 5 (5.7%) | 1 (0.5%) |
| HRT | 9 (10.2%) | 18 (9.0%) |
| Oral contraceptive pill | 0 (0.0%) | 5 (2.5%) |
| **Smoking status, n (%)** | | |
| Smoker | 59 (18.7%) | 43 (11.5%) |
| Ex-smoker | 126 (40.0%) | 143 (38.2%) |
| Non-smoker | 130 (41.3%) | 188 (50.3%) |
| **Alcohol drinking status, n (%)** | | |
| Drinker | 271 (86.3%) | 347 (92.8%) |
| Ex-drinker | 21 (6.7%) | 10 (2.7%) |
| Non-drinker | 22 (7.0%) | 17 (4.5%) |
| **Laboratory examinations, median (IQR)** | | |
| Glucose, mmol·L$^{-1}$ | 5.64 (2.08) | 5.11 (1.04) |
| SBP, mmHg | 146 (21) | 142 (19) |
| DBP, mmHg | 84 (11) | 83 (11) |
| Cholesterol, mmol·L$^{-1}$ | 5.71 (1.47) | 5.82 (1.11) |
| BMI, kg·m$^{-2}$ | 28.36 (4.67) | 27.32 (4.32) |

[a]Abbreviations: MI, myocardial infarction; HRT, hormone replacement therapy; SBP, systolic blood pressure; DBP, diastolic blood pressure; HDL, high-density lipoprotein; BMI, body mass index. IQR, interquartile range.

### Performance of GPTs

When provided with complete patient descriptions, the overall performance of ChatGPT in the prediction of myocardial infarction is not as well as expected, with AUC of 0.62 (95% CI: 0.58~0.66, Figure 1), only marginally better than the random chance. GPT-4, a more advanced version, is significant outperformed than that of ChatGPT, with AUC of 0.69 (95% CI: 0.65 – 0.73, Figure 1), suggesting that updated model architecture have enhanced its ability to accurately predict myocardial infarction.

Under optimized threshold, ChatGPT predicts MI with accuracy of 0.59 (95% CI: 0.55~0.63, Table S1), similar to the indices provided by Jee (0.581, 95% CI: 0.579~0.583, Table S3) and Wilson (0.597, 95% CI: 0.596~0.599, Table S3), yet significantly lower than that of GPT-4 (0.65, 95% CI: 0.62~0.69, Table S2). GPT-4 shows similar sensitivity to the indices, all significantly higher that of ChatGPT, yet its specificity was the lowest in all models (see Figure 4). Therefore, both ChatGPT and GPT-4 are limited in predicting MI incidence, while GPT-4 was relatively more sensitive. Together, ChatGPT and GPT-4 shows a weak performance that is merely better than "guessing" the result in predicting MI with problems probably due to model architecture rather than data.

### Performance of GPTs under Chain of Thought

To test whether the predictions of ChatGPT and GPT-4 are generated logically, an experiment was designed using CoT questioning. The comparison between ChatGPT and GPT-4 for myocardial infarction prediction reveals clear differences in performance across both models. The CoT approach, where information was progressively added through a series of questions, resulted in lower AUCs compared with that given complete information. In the case of ChatGPT, CoT groups has AUCs between 0.56 and 0.58 (Group 1: 0.56, 95% CI: 0.52~0.61; Group 2: 0.56, 95% CI: 0.52~0.60; Group 3: 0.58, 95% CI: 0.54~0.62, Figure 2A, Table S1), and GPT-4 witnesses slightly better but still modest improvements, with AUCs between 0.60 and 0.61(Group 1: 0.61, 95% CI: 0.57~0.65; Group 2: 0.60, 95% CI: 0.56~0.64; Group 3: 0.60, 95% CI: 0.56~0.65, Figure 2B, Table S1). Therefore, CoT questioning approach yields no substantial improvements in predictive accuracy, indicating that simply expanding the scope of the inquiry in a stepwise fashion does not result in enhanced model accuracy. The direct use of all available information in one query outperforms the CoT method, particularly for GPT-4, suggesting less logical in the prediction process of ChatGPT and GPT-4.

### Model Comparisons

Finally, we compared GPT models (GPT-4 and ChatGPT) with traditional machine learning models (e.g., decision tree, logistic regression, random forest and SVM), other mainstream NLP models (e.g., Transformer, BERT), as well as open-source large language models (LLMs) like Llama, Mixtral, MedLlama, and Meditron. The following provides a detailed comparison of their performance in predicting myocardial infarction.

### GPT Models vs Traditional Machine Learning

We compared GPTs with existing machine learning models. AUCs of ChatGPT and GPT4 are significance lower than that of machine learning models (Decision tree: 0.79, 95% CI: 0.76~0.83; Logistic regression: 0.77, 95% CI: 074~0.81; Random Forest: 0.76, 95% CI: 0.73~0.80; SVM: 0.79, 95% CI 0.76~0.83, Figure 3A). Under optimized threshold, GPTs is predicting less accurate than machine learning models (Decision tree: 0.78, 95% CI: 0.74~0.81; Logistic regression: 0.77, 95% CI: 0.73~0.80; Random Forest: 0.77, 95% CI: 0.74~0.80; SVM: 0.78, 95% CI: 0.75~0.81, Figure 4) with significant discrimination (Figure 4, Table S4). This result indicates that data-driven machine learning models such as SVM and decision tree have achieved a certain level of clinical diagnostic capability. In contrast, GPTs, especially ChatGPT, only outperform random guessing in overall accuracy and exhibits a significant difference compared to machine learning models trained on external datasets.

### GPT Models vs. Traditional NLP Models

GPT-4 with compete information achieves the highest AUC of 0.69 (95% CI: $0.65 - 0.73$), showing superior performance compared to all other models in this set. ChatGPT with complete information has a moderate AUC of 0.62 (95% CI: $0.58 - 0.66$), outperforming traditional NLP models but falling short of GPT-4. BERT achieved an AUC of 0.57 (95% CI: $0.53 - 0.61$, Figure 3B, Table S5), while the Transformer model underperformed with an AUC of 0.51 (95% CI: $0.47 - 0.55$, Figure 3B, Table S5), approaching random prediction. This indicates that the GPT models, especially GPT-4, are better equipped for tasks requiring deeper understanding and reasoning, outperforming traditional NLP models like BERT and Transformer, which may struggle with complex predictive tasks due to their simpler architecture and training on general language tasks.

### GPT Models vs. Open-Source Large Language Models

To confirm whether the performance of GPTs represent common features of large language models, other open source LLMs were also tested. GPT-4 with complete information again shows the best performance with an AUC of 0.69 (95% CI: $0.65 - 0.73$), indicating a notable lead over other large models. ChatGPT with complete information achieves an AUC of 0.62 (95% CI: 0.58−0.66), similar to its performance in the previous comparison. Llama, an open-source general-purpose model, demonstrates a competitive AUC of 0.61 (95% CI: 0.56 − 0.65, Figure 3C). MedLlama, a fine-tuned version of LLAMA on medical data, shows slight improvement with an AUC of 0.62 (95% CI: 0.57 − 0.66, Figure 3D), suggesting that domain-specific fine-tuning brings a modest benefit. Mixtral (Figure 3C) and Meditron (Figure 3D) both have an AUC of 0.60 (95% CI: 0.56 − 0.65), reflecting decent performance, but still falling short of GPT-4 (Figure 4, Table S6). These results show that while open-source models such as LLAMA, MedLlama, and Meditron perform reasonably well, GPT-4 demonstrates a clear advantage, especially when applied to medical tasks. The improvements seen in MedLlama, through domain-specific fine-tuning, indicate that medical adaptation of models improves their predictive capability, though it still does not surpass the general-purpose GPT-4.

Across both comparisons, GPT-4 consistently outperforms traditional NLP models like BERT and Transformer, as well as large open-source models like Llama and its medically fine-tuned counterparts (Figure 4). This highlights strong capability of GPT-4 in handling complex prediction tasks in the medical domain, even without specific medical fine-tuning. The fine-tuning of models on medical datasets, as seen with MedLlama and Meditron, offers noticeable but not groundbreaking improvements. Fine-tuning these models brings their performance closer to general-purpose GPT models, but they do not yet surpass the advanced architectures like GPT-4, which have been trained on significantly larger and more diverse datasets.

In conclusion, existing open-source large language models (LLMs) such as Llama, MedLlama, and Meditron still lack the necessary precision for clinical diagnosis, with their performance falling short of the requirements for reliable medical decision-making. While GPT-4 achieves the best results among all models, its accuracy remains limited, with an AUC of 0.69, which is still below the threshold needed for practical clinical applications.

## Discussion

In this retrospective cohort study based on UK Biobank database, the predictability of LLMs represented by ChatGPT and GPT-4 on MI, a common disease, has been evaluated and quantified. Given the importance of early diagnosis of MI, we set up a pre-clinical scenario to imitate real-world chats where individualized data were transformed into prompts for the chatbots. Despite successful application of LLMs on NLP tasks of clinical medicine, the ability of LLMs in quantified task is merely as similar to simple medical indices with no manifested capabilities for conjecturing logical inference. It seems that current LLMs predominantly designed for text recognition and pattern generation tasks lack robust and quantified logical reasoning capabilities, which are crucial for making complex clinical decisions.

Basically, ChatGPT is a representative large-scale language model with few-shot in-context prompt learning ability primarily designed for natural language conversations [23]. These LLMs are able to make flexible interactions, dynamic text specification and multi-model inputs with medical domain knowledge and potential in bedside decision support [9]. Despite its enormous capabilities in natural language generation, ChatGPT may exhibit limitations when applied to real-world tasks within professional domains [24]. It has been found that GPT-3, a previous version of ChatGPT, provided diagnosis below physicians and better than lay individuals, but closer to the latter [25]. Also, other LLMs such as Bing Chat and Bard showed limitations in answering professional dentistry questions, even returning harmful results [26]. Reasons for this phenomenon may include the lack of specific domain expertise, the broad and uncertain sources of training data, contextual understanding challenges and the lack of control over dialogue generation. Therefore, LLMs expert in medical domain may be developed in the future, either by fine-tuning or prompt-tuning based on medical data, or applying in-context learning to understand medical texts.

In addition, the CoT approach, designed to simulate human-like stepwise reasoning, failed to leverage the incremental data in a way that improved prediction, suggesting a lack of logical reasoning ability in current LLMs similar to that observed in previous study [27]. In principle, the CoT method mimics how humans might think: starting with basic information and adding complexity in stages to make more refined decisions. While GPT models excel in pattern recognition across large datasets, they do not naturally infer causality or establish deep connections between variables when new data is added progressively. Logical thinking requires an understanding of the background, knowledge, and context of a problem, as well as the ability to engage in reasoning, judgment, and induction. The answers generated by ChatGPT, however, are based on the training dataset it has been provided with, not developing the ability to engage in logical thinking and accurately understand the context of a question [28]. GPT models do not naturally form such logical chains, as they lack a structured reasoning process akin to human cognition.

Despite their failure in predicting MI logically, future LLMs are still expected for clinical medicine, including CDS tasks. For LLMs extract natural language information rather than structural medical datasets in tabular forms, these models can be used to deal with unstructured text-based medical data such as medical records. More importantly, clinical scientific rules proved by evidence-based medicine might become an indispensable element in future LLM-based CDS systems. High-quality information based on scientific researches and empirical clinical knowledge may promote further development of medical LLMs, where information given by natural languages would be encoded, extracted, characterized and denoised, and later on combined deeply with updated scientific rules to make the most appropriate decisions [29].

### Limitations

This study has limitations. Firstly, the diagnosis of MI is "the earlier the best" [7], yet the time window of ChatGPT was selected as one year in the current study due to data limitations. Secondly, the diagnosis of MI usually requires an initial judgment of physicians, followed by in-depth examinations [7]. The current study focuses on pre-admission prediction, but clinically, this dynamic process should also be considered in evaluating the performance of LLMs. We also recommend that noise be appropriately added into the questioning session to validate the robustness of the results.

### Conclusions

This study has set out to make quantified evaluation and comparative analysis on the predictive value of LLMs in a common CDS task based on a large-scale clinical cohort. LLMs represented by ChatGPT and GPT-4 are currently weak in making prediction logically with medical knowledge inferences, and thus are still not ready to be applied in clinical medicine fields. However, future medical LLMs are suggested to develop large models that are expert in medical domain knowledge with the ability to understand both natural languages and quantified medical data, and make logical inferences towards the most accurate decision.

### Acknowledgements

We deeply appreciate the insightful comments provided by reviewer J and reviewer DJ, and these comments greatly helped to enhance the article.

Y. D and G. H contributed to conceptualization, funding acquisition, writing – review & editing; Y. Z, Y. G and K. Y conceived to methodology, formal analysis and writing – original draft; A.C. Y contributed to resources and data curation; H. W, H. X and H. Y contributed to validation. All authors read and approved the final manuscript.

This work was supported by the National Key Research and Development Project (2019YFC1710104), the National Natural Science Foundation of China (grant number: 91942301). The funder played no role in study design, data collection, analysis and interpretation of data, or the writing of this manuscript.

The underlying code for this study is not publicly available but may be made available to qualified researchers on reasonable request from the corresponding author.

**Conflicts of Interest**

All authors declare no financial or non-financial competing interests.

**Abbreviations**

LLM: large language model
NLP: natural language processing
CDS: clinical decision support
SVM: support vector machine

# Supplementary Appendix

## SUPPLEMENTARY METHODS
### eMethod 1: Data Filtering and Resampling
*Inclusion/Exclusion Criteria*

A total of 502,468 participants were first filtered under the inclusion/exclusion criteria. Subjects would be included if they meet with the following criteria:

(1) Without MI incidence at baseline;

***AND***

(2) Recruitment within one year (365 days) before MI incidence;

***AND***

(3) Without missing data on basic information: i.e., age and gender.

Participants meet with the following criteria would be excluded to guard the quality of the dataset:

(1) Lost of follow-up;

***OR***

(2) Dead at baseline.

*Resampling*

A primary cohort have been built with UK Biobank participants filtered by the above inclusion/exclusion criteria, which turns out to be enough for the current study. However, some issues exist in the primary which might influence the result of the study.

The first issue is the unbalanced ratio of participants. Only 0.14% of participants were recorded to experience myocardial infarction incidence within the observation duration. This brings the probability that despite the model might predict all of the participants to have no myocardial infarction, it would still achieve an accuracy of 99.86%. Therefore, the primary cohort was further resampled to a secondary cohort to balance the proportion.

The second issue is the rate limit set by OpenAI API measured by RPM (requests per minute) and TPM (tokens per minute). Only 3 RPM and 40,000 TPM was allowed for free-trial users and 60 RPM and 60,000 TPM for Pay-as-you-go users when using the model *gpt-3.5-turbo*, generally known as ChatGPT. This limit also requires a smaller cohort to be constructed.

The final cohort was resampled from the primary cohort, in which sample size was estimated restrictedly. Index of participants were selected by random number generated by a given seed using Python.

*Resample Size Estimation*

The sample size of the final cohort was estimated based on sensitivity and specificity according to the formula widely used for diagnostic trials in epidemiology as follows:

$$n = \frac{u_\alpha^2 \times p \times (1-p)}{\delta^2}$$

where,

$n$ denotes sample size in each single group,

$p$ denotes the probability in each group,

$u_\alpha$ is the quantile under significant level $\alpha$,

$\delta$ represents absolute precision desired.

We used previous results as pre-experiment, where ChatGPT performed a sensitivity of 71% and a specificity of 42%. Under significant level $\alpha = 0.05$ and precision level $\delta = 0.05$, a total of 316 participants were needed to be re-sampled from those with MI incidence, and a total of 374 participants from those without MI incidence. The proportion between MI and non-MI is thus 316/374=0.84. Therefore, 790 participants were finally sampled randomly.

**eMethod 2: Prompt Generalization and ChatGPT Response**

According to the selected risk factors for myocardial infarction, we divided the survey data into three groups and applied a chain-of-thought prompting approach. The first group focuses on the participant's basic information, symptoms, and medical history, establishing a foundational understanding. Building on this, the second group includes personal history, expanding the scope of inquiry. The third group further deepens the analysis by incorporating family history, completing the logical progression of information gathering.

Secondly, we converted the tabular data of the participants provided by the UK Biobank into self-descriptions of the interviewed participants. Specifically, we simulated the participants describing their own interview situation in the first-person narrative form during a medical visit. For example, a 41-year-old White male with symptoms of "wheezed or whistled in the chest in the last year", blood glucose level of 7.812 mmol/L, diastolic blood pressure of 84 mmHg, systolic blood pressure of 144 mmHg, cholesterol level of 4.023 mmol/L, and a BMI of 21.212 kg/m$^2$. We transformed the above information into the following question format to ask ChatGPT, following the relevant indicators defined by international medicine: "I am a 41-year-old White male. I wheezed or whistled in the chest in the last year. I have diabetes, with a blood glucose level of 7.812 mmol/L. I have hypertension, with blood pressure of 144/84 mmHg. I do not have hypercholesterolemia. I do not have obesity." An example of the question format is shown in Figure S1.

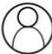

**Figure S1. One example of ChatGPT questioning.**

Finally, we asked ChatGPT to select a discrete value from 1 to 10 that it considers to be the closest estimate of the current risk factor of the interviewed participant having myocardial infarction. Other cases in ChatGPT questioning are shown in Figure S2.

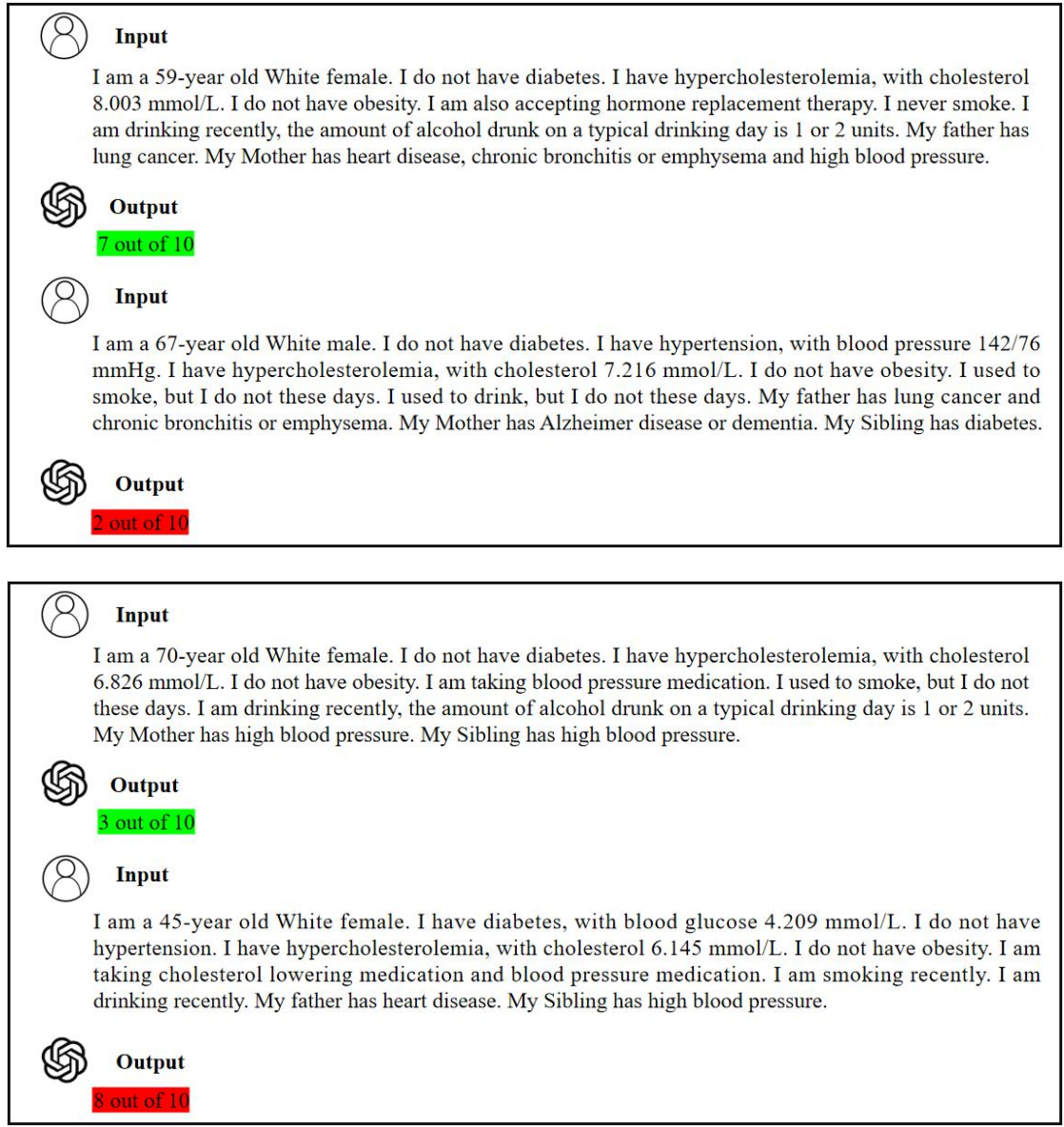

**Figure S2. Other cases in ChatGPT questioning.**

The true labels for the first two inputs are diagnosed with myocardial infarction, while the true labels for the last two inputs are not having myocardial infarction. The answers marked in green represent correct predictions, while those marked in red represent incorrect predictions.

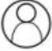

**Figure S3. one cases in ChatGPT questioning with Chain of thought.**

The design of the chain-of-thought questioning technique is structured to promote a logical and sequential exploration of complex topics by breaking down the question into a series of interconnected sub-questions. This method typically begins with broad, foundational inquiries and gradually narrows down to more specific or detailed factors. In constructing the chain, the initial questions often address general variables (e.g., age, lifestyle) that set the context. Subsequent questions introduce more specific considerations (e.g., family history, medical conditions), creating a cumulative framework of thought.

**eMethod 3: Model Comparison**
*Equation-Based Prediction Models (Medical indices)*

Through selection, we applied two comparative models that included different risk factor indicators, as well as gender, race, and other information.

Wilson et al. developed gender-specific prediction equations for coronary heart disease (CHD) risk based on factors such as age, diabetes, smoking, JNC-V blood pressure category, and cholesterol. Jee et al. developed a CHD prediction model for Koreans and compared the estimated values of the model with actual CHD cases.

*NLP Models*

Transformer is a deep learning model architecture used for natural language processing (NLP) tasks, proposed by Vaswani et al. in 2017. It introduces the concept of self-attention mechanism to capture contextual information in input sequences and achieves parallel computation, replacing RNN as the main architecture of current mainstream NLP models.

Bert is a pre-trained language model based on the transformer architecture. It was introduced by Google in 2018 and utilizes a bidirectional context modeling approach. Traditional language models usually use unidirectional models from left to right or right to left, while BERT processes input sequences in a bidirectional manner using transformer encoders to better capture contextual information. This bidirectional modeling approach enables Bert to have better semantic understanding and representation capability when dealing with natural language tasks. The advantage of BERT is its ability to leverage a large amount of unlabeled data for pre-training, learning general language representations. Through fine-tuning, the parameters of the BERT model are further optimized, resulting in excellent performance on various downstream tasks.

*Other LLMs*

LLAMA is a series of large-scale autoregressive language models developed by Meta. LLAMA is designed to be highly efficient in terms of computational resources while achieving competitive performance compared to much larger models. It emphasizes scaling up model size while maintaining cost-effective training and inference through optimizations such as more effective parameter sharing and fine-tuned training techniques. LLAMA models are typically utilized for tasks in natural language understanding and generation, including question answering, summarization, and translation, and they have set benchmarks in terms of both accuracy and computational efficiency across various NLP tasks.

Mixtral is a specialized multilingual language model developed with a focus on low-resource languages. It leverages a mixture of translation data, linguistic knowledge, and pretraining on a wide variety of languages, both high-resource and low-resource. Mixtral's architecture allows for effective cross-lingual transfer, making it particularly strong in scenarios where language resources are sparse. By employing techniques such as language embeddings and modular parameter sharing, Mixtral aims to balance model performance across diverse languages while maintaining high performance in language-specific tasks, including machine translation, speech recognition, and language modeling in underrepresented languages.

MedLlama is a domain-specific variant of the LLAMA model, tailored for medical and clinical applications. It has been fine-tuned on extensive biomedical corpora, including medical literature, clinical reports, and structured healthcare datasets. MedLlama's architecture allows it to excel in tasks requiring specialized medical knowledge, such as clinical decision support, medical information retrieval, and automated diagnosis generation. By incorporating biomedical terminologies and leveraging context-aware learning, MedLlama provides higher accuracy and relevance in medical dialogue systems, clinical note summarization, and other healthcare-related natural language processing (NLP) tasks.

Meditron is an advanced transformer-based language model specifically designed for healthcare-related natural language processing tasks. It has been trained on vast corpora of medical literature, including diagnostic reports, research papers, and healthcare guidelines. Meditron is structured to support critical medical applications such as automated medical coding, patient record analysis, and clinical question answering. The model integrates domain-specific knowledge with transformer-based architectures to enable a deep understanding of medical jargon, abbreviations, and complex biomedical language, making it a powerful tool for clinicians, researchers, and health informatics systems.

*LLMs fine-tuning strategy*

In this study, we used all datasets excluding the sampling part for fine-tuning. During training, the data was split 8:2 into training and validation sets. After fine-tuning, we tested the model on the sampled test, allowing us to evaluate its performance. We then compared its results on the same sampled test set with those obtained using ChatGPT and other methods, ensuring a consistent and fair comparison between models. We carried out The fine-tuning process on severe models. To ensure the validity of the model training, we maintained the original proportion of non-MI and MI samples in both the training and validation datasets. Each of the models was initialized using the pre-trained weights available from Hugging Face, ensuring that the models retained their previously learned knowledge from large-scale pre-training. To adapt these models for binary prediction tasks, we added a fully connected layer on top of the final output layer of each model. This additional layer was designed specifically for binary classification, consisting of a single neuron with a sigmoid activation function to output probabilities for the two classes (e.g., 0 or 1). We used the binary cross-entropy loss function (BCE) to optimize the models. We present the versions of other large language models used in the study, along with the specific training strategies, as outlined below:

1. Transformer Model: The Adam optimizer was used with an initial learning rate of 3e-5, a batch size of 64, and the number of training epochs set 50.

2. BERT Base Model: The Adam optimizer was used with an initial learning rate of 2e-5, a batch size of 64, and the number of training epochs set 50. To prevent overfitting, a dropout rate of 0.1 was applied during training.

3. Meta-Llama-3.1-8B-Instruct: The model, with 8 billion parameters, was fine-tuned using a learning rate of 1e-5, a batch size of 32, and mixed-precision training for efficiency. The training was conducted for 50 epochs, with dynamic learning rate adjustments based on validation performance.

4. JSL-MedLlama-3-8B-v2.0-GGUF: The model was fine-tuned using an AdamW optimizer with a learning rate of 1.5e-5 and gradient clipping to avoid exploding gradients. Mixed-precision training was employed to accelerate the process, with batch sizes of 24, and training was conducted for 50 epochs.

5. Meditron-7B: We employed the Adam optimizer with weight decay, a learning rate of 3e-5, and a batch size of 48. The model was fine-tuned over 50 epochs, and knowledge distillation techniques were used to improve the fine-tuned model's performance through knowledge transfer from a smaller model.

6. Mixtral-8x7B-Instruct-v0.1: The model was fine-tuned with a learning rate of 2e-5, a batch size of 36, and trained for 50 epochs. Mixed-precision training was employed to improve efficiency. Additionally, gradient accumulation was used to handle larger batches and improve convergence.

## SUPPLEMENTARY RESULTS

Table S1. Performance of ChatGPT.

| Groups  | AUC              | Ac               | Se               | Sp               | YI               |
|---------|------------------|------------------|------------------|------------------|------------------|
| All     | 0.62 (0.58~0.66) | 0.59 (0.55~0.63) | 0.45 (0.40~0.50) | 0.76 (0.71~0.80) | 0.21 (0.11~0.31) |
| Group 1 | 0.56 (0.52~0.61) | 0.56 (0.52~0.60) | 0.52 (0.46~0.57) | 0.62 (0.56~0.67) | 0.13 (0.03~0.24) |
| Group 2 | 0.56 (0.52~0.60) | 0.53 (0.50~0.57) | 0.35 (0.30~0.40) | 0.75 (0.70~0.80) | 0.10 (0.00~0.20) |
| Group 3 | 0.58 (0.54~0.62) | 0.55 (0.52~0.59) | 0.32 (0.27~0.37) | 0.84 (0.79~0.87) | 0.15 (0.06~0.24) |

Abbreviations: AUC, area under curve; Ac, accuracy; Se, sensitivity; Sp, specificity; PV, predictive value.

**Table S2. Performance of GPT-4.**

| Groups | AUC | Ac | Se | Sp | YI |
|---|---|---|---|---|---|
| All | 0.69 (0.65~0.73) | 0.65 (0.62~0.69) | 0.62 (0.57~0.67) | 0.70 (0.64~0.75) | 0.31 (0.21~0.41) |
| Group 1 | 0.61 (0.57~0.65) | 0.59 (0.56~0.63) | 0.61 (0.56~0.66) | 0.58 (0.52~0.63) | 0.18 (0.07~0.29) |
| Group 2 | 0.60 (0.56~0.64) | 0.59 (0.55~0.62) | 0.68 (0.63~0.73) | 0.47 (0.42~0.53) | 0.16 (0.05~0.26) |
| Group 3 | 0.60 (0.56~0.65) | 0.59 (0.55~0.63) | 0.60 (0.55~0.65) | 0.57 (0.52~0.63) | 0.17 (0.07~0.28) |

Abbreviations: AUC, area under curve; Ac, accuracy; Se, sensitivity; Sp, specificity; PV, predictive value.

**Table S3. Performance of medical indices.**

| Indices | AUC | Ac | Se | Sp | YI |
|---|---|---|---|---|---|
| Jee | 0.74 (0.71~0.76) | 0.581 (0.579~0.583) | 0.581 (0.579~0.583) | 0.80 (0.76~0.84) | 0.38 (0.34~0.42) |
| Wilson | 0.73 (0.70~0.75) | 0.597 (0.596~0.599) | 0.597 (0.596~0.599) | 0.76 (0.71~0.80) | 0.35 (0.31~0.40) |

Abbreviations: AUC, area under curve; Ac, accuracy; Se, sensitivity; Sp, specificity; PV, predictive value.

**Table S4. Performance of traditional machine learning models.**

| Models | AUC | Ac | Se | Sp | YI |
|---|---|---|---|---|---|
| Decision Tree | 0.79 (0.76~0.83) | 0.78 (0.74~0.81) | 0.76 (0.71~0.80) | 0.80 (0.75~0.84) | 0.56 (0.46~0.64) |
| SVM | 0.79 (0.76~0.83) | 0.78 (0.75~0.81) | 0.79 (0.75~0.83) | 0.77 (0.72~0.82) | 0.56 (0.47~0.65) |
| Logistic | 0.77 (0.74~0.81) | 0.77 (0.73~0.80) | 0.77 (0.72~0.81) | 0.77 (0.72~0.81) | 0.64 (0.44~0.62) |
| Random Forest | 0.76 (0.73~0.80) | 0.77 (0.74~0.80) | 0.78 (0.73~0.82) | 0.77 (0.72~0.81) | 0.55 (0.45~0.63) |

Abbreviations: AUC, area under curve; Ac, accuracy; Se, sensitivity; Sp, specificity; PV, predictive value.

**Table S5. Performance of traditional natural language processing models.**

| Models | AUC | Ac | Se | Sp | YI |
|---|---|---|---|---|---|
| Transformer | 0.51 (0.47~0.55) | 0.50 (0.46~0.54) | 0.32 (0.27~0.37) | 0.72 (0.67~0.77) | 0.04 (-0.06~0.14) |
| BERT | 0.57 (0.53~0.61) | 0.57 (0.53~0.61) | 0.51 (0.46~0.56) | 0.64 (0.58~0.69) | 0.14 (0.04~0.25) |

Abbreviations: AUC, area under curve; Ac, accuracy; Se, sensitivity; Sp, specificity; PV, predictive value.

**Table S6. Performance of other LLMs.**

| Models | AUC | Ac | Se | Sp | YI |
|---|---|---|---|---|---|
| Llama | 0.61 (0.56~0.65) | 0.60 (0.57~0.64) | 0.60 (0.55~0.65) | 0.60 (0.55~0.66) | 0.21 (0.10~0.31) |
| Mixtral | 0.60 (0.56~0.65) | 0.62 (0.58~0.65) | 0.63 (0.58~0.68) | 0.60 (0.55~0.66) | 0.23 (0.13~0.34) |
| MedLlama | 0.63 (0.57~0.66) | 0.63 (0.59~0.67) | 0.63 (0.58~0.68) | 0.63 (0.57~0.68) | 0.26 (0.16~0.37) |
| Meditron | 0.60 (0.56~0.65) | 0.61 (0.58~0.65) | 0.62 (0.57~0.67) | 0.61 (0.55~0.66) | 0.23 (0.12~0.33) |

Abbreviations: AUC, area under curve; Ac, accuracy; Se, sensitivity; Sp, specificity; PV, predictive value.